\documentclass{article}
\usepackage{spconf,amsmath,graphicx}
\usepackage{xcolor}
\usepackage{soul}

\title{D2SLAM: Semantic visual SLAM based on the Depth-related influence on object interactions for Dynamic environments}
\twoauthors
  {Ayman Beghdadi, Malik Mallem}
	{IBISC Lab, University Paris-Saclay\\
Computer Engineering Department\\
Evry, France}
  {Lotfi Beji}
	{IBISC Lab, University Paris-Saclay\\
Mechanical Engineering Department\\
Evry, France}
\begin{document}
%\ninept
%
\maketitle
\begin{abstract}
Considering the scene's dynamics is the most effective solution to obtain an accurate perception of unknown environments for real vSLAM applications. Most existing methods attempt to address the non-rigid scene assumption by combining geometric and semantic approaches to determine dynamic elements that lack generalization and scene awareness. We propose a novel approach that overcomes these limitations by using scene-depth information to improve the accuracy of the localization from geometric and semantic modules. In addition, we use depth information to determine an area of influence of dynamic objects through an Object Interaction Module that estimates the state of both non-matched and non-segmented key points. The obtained results on TUM-RGBD dataset clearly demonstrate that the proposed method outperforms the state-of-the-art.
\end{abstract}
\begin{keywords}
Depth influence, Dynamic environments, Robust, Semantic segmentation, Visual SLAM 
\end{keywords}
\section{Introduction}
\label{sec:intro}

Visual Simultaneous Localization And Mapping (vSLAM) has been widely investigated over the past decade for deploying many applications in robotics \cite{ORB2,SVO,VINS}. 
This technology, based on extracting visual information, attempts to locate the robot in an unknown environment. 
Its deployment in real world applications has highlighted their sensitivity to the dynamics of the scene \cite{saputra2018visual,beghdadi2022comprehensive}.
Indeed, the vSLAM attempts to locate the camera in its environment using visual landmarks tracked during the camera's motion. 
%Thus, moving visual landmarks produce an error in the localization estimation, affecting the precision. 
Thus, moving visual landmarks induce errors in the localization estimation process, thus affecting its accuracy.
To solve this problem, many methods \cite{Dyna,Dyn,DS,DP} deal with the dynamics through two approaches by classifying the keypoints into static and dynamic states to reject those in a dynamic state from the vSLAM process.
First, the semantic segmentation approach determines the dynamic objects according to their nature and the type of environment by segmenting the scene, where only humans are dynamic in indoor environments. This approach is so limited to object classes that the model can segment.
Subsequently, the geometric approach \cite{hartley2003multiple} was introduced to consider the whole image and deal with the other moving objects. It estimates the reprojection error of matched keypoints of two successive frames using the epipolar geometry constraints.
Unfortunately, this approach has a limited impact since it only determines the state of matched keypoints.
%Besides, geometric reasoning does not consider 3D information from the depth, which may induce errors in estimating the state of keypoints. 
Besides, geometric reasoning does not consider depth information. This may induce errors in estimating the state of keypoints.
Indeed, according to the projective geometry, the displacement of a distant point induces a smaller reprojection error than a closer moving point with the same motion amplitude. 
Likewise, the semantic approach supposes that predicted masks have constant reliability. 
However, many works have shown that the depth through the object's size affects the segmentation accuracy.
To overcome these limitations, we propose a novel approach based on the ORB-SLAM3 framework \cite{ORB} by considering the depth information. It allows to improve the keypoints state estimates through the geometric and semantic modules and those that are neither matched nor in a segmented area.
The keypoints state estimates provided by the two modules are then refined thanks to an adaptive depth-related thresholding process.
In addition, we assume that dynamics result only from human activity in indoor environments. Thereby, the probability that static objects become dynamic increases if they are within the zone of human influence, i.e., at a depth and a 2D image position (2D-plus-Depth) close to humans. Thus, we estimate the state of non-matched and non-segmented keypoints by analyzing their neighborhood in these zones. Matched keypoints, considered as dynamic through the geometric module, inform us about the probability of the state of unpaired keypoints without requiring additional semantic information from classes other than humans.
\begin{figure*}[t]
    \centering
        \includegraphics[width=0.93\textwidth]{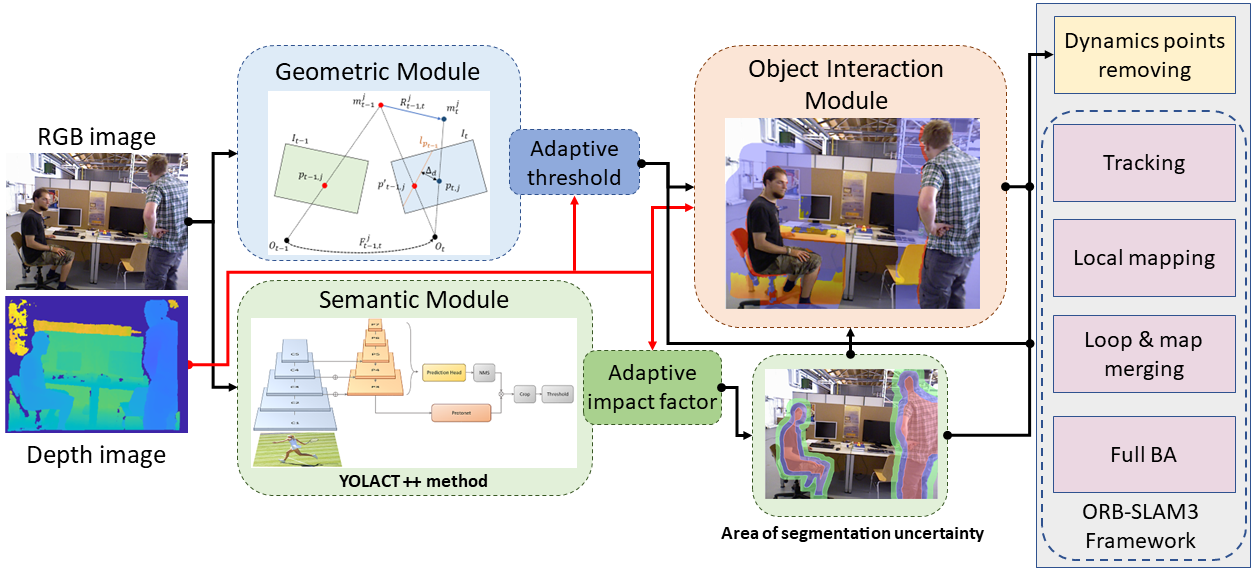}
        \caption{Architecture of our proposed D2SLAM method}
    \label{system}
\end{figure*}
The main contributions and novelty of this work are summarized as follows:
\begin{itemize}
\vspace{-0.25em}
    \item A new efficient method dealing with scene dynamics outperforming the state-of-the-art on the dedicated TUM-RGBD dataset is proposed.
    \vspace{-0.25em}
    \item A refinement process of keypoint state estimates through probabilistic functions using an adaptive depth-related thresholding and impact factor in the geometric and semantic modules, respectively, is introduced.
    \vspace{-0.25em}
    \item An Object Interaction Module (OIM), that estimates the non-matched and non-segmented keypoints state considering the area of influence of dynamic objects without requiring extra semantic classes, is introduced.  
\end{itemize}
The paper is organized as follows. Section \ref{work} summarizes the related works. Section \ref{method} is devoted to describe our method. The obtained results are presented and discussed in section \ref{Expe}. Finally, concluding remarks are provided in section \ref{conclusion}.
%The paper is organized as follows. Section \ref{work} summarizes the related works. Sections \ref{method} and \ref{Expe} describe the method and results. Finally, section \ref{conclusion} provides concluding remarks.

\section{Related Work}
\label{work}

% In this section, we present related works on vSLAM methods that attempt to deal with dynamic environments.
%through the geometric and semantic segmentation approaches. 
%Most of them are based on the ORB-SLAM2 or ORB-SLAM 3 frameworks and include deep learning networks.
Existing vSLAM methods dealing with dynamic environments such as DynaSLAM \cite{Dyna} associates the multi-view geometry, without considering the depth-impact and semantic segmentation information to detect moving objects. 
DS-SLAM \cite{DS} determines dynamics by checking the moving consistency through the epipolar geometry, which is combined with a segmentation module requiring extra semantic information (desk, chair, etc.).
Similarly, Dynamic-SLAM \cite{Dyn} requires extra semantic information and prior knowledge for a reliable object detection method to detect dynamic keypoints in the scene.
DP SLAM \cite{DP} is based on estimating the moving probability propagation of the dynamic keypoints combining the epipolar geometry constraints and semantic segmentation into a Bayesian filter. This method does not require extra information but does not consider the depth information.
DGS-SLAM \cite{DGS} uses a multinomial residual network to detect dynamic objects combining the motion information from consecutive frames and potential motion information from the semantic segmentation. 
%It achieves a robust camera pose tracking strategy using a keypoints classification. A segmentation module employs these modules to extract potential moving points thanks to a semantic frame selection strategy.  
This method estimates the state of the uncertain keypoints by considering the global depth via the K-mean Clustering algorithm. However, it does not refine this estimate by analysing the neighborhood of these keypoints. This may result in misclassification of the keypoint states.

\section{Proposed method}
\label{method}

The architecture of our vSLAM method is illustrated in fig. \ref{system}. It consists of three modules, namely, the geometric, the semantic, and the object interaction module.
The segmentation, based on the YOLACT++ \cite{bolya2020yolact++} method, only generates masks of humans present in the scene without using any extra semantic information. 
%\vspace{-0.5em}
\subsection{Semantic segmentation approach}\label{AA}
The keypoints state of the semantic module is obtained by calculating their displacement probability related to their distance from the mask edges. First, the minimal distance $\Delta d_{m}$ between a keypoint $p_{i}$ and a point $m_{i}^{n}$ on the edge of the mask $n$ is computed as:
\begin{equation}
\Delta d_{m} = \min \Vert p_{i} - m_{i}^{n} \Vert  \label{eq1}
\end{equation}
This distance is used as a key factor in a binomial logistic regression model for estimating the probability of the keypoints state, according to the mask $n$, as follows:
\begin{equation}
P(S_{p_{i}^{n}}) = \frac{1}{exp(- \beta(z_{p_{i}}) \cdot \Delta d_{m})+1}    \label{eq2}
\end{equation}
\begin{figure}[h]
    \centering
        \includegraphics[width=0.48\textwidth]{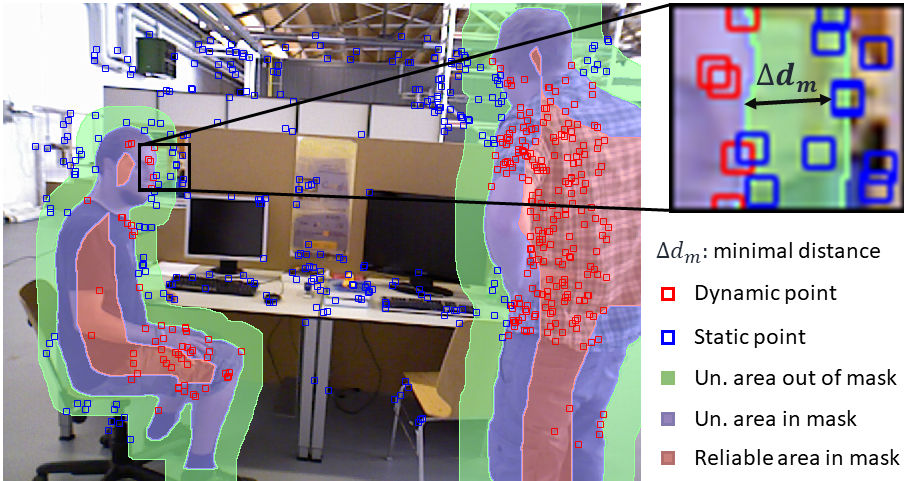}
        \caption{Areas of segmentation uncertainty}
        \vspace{-0.5em}
    \label{mask}
\end{figure}
Where $\beta(z_{pi}) \in [0.05;0.25]$ is the adaptive impact factor proportional to the depth $z_{p_{i}}$ of the keypoint $p_{i}$. 
This factor implicitly defines the uncertainty area inside and outside the mask  as described in fig.\ref{mask}. 
Therefore, the greater the distance, the more accurate the estimation of the state of this point as static (outside the mask) or dynamic (inside the mask). 
This factor is used to determine the mask uncertainty zone corresponding to a state probability lower than $75\%$.

\subsection{Geometric approach}
\begin{figure}[h]
    \centering
        \includegraphics[width=0.35\textwidth]{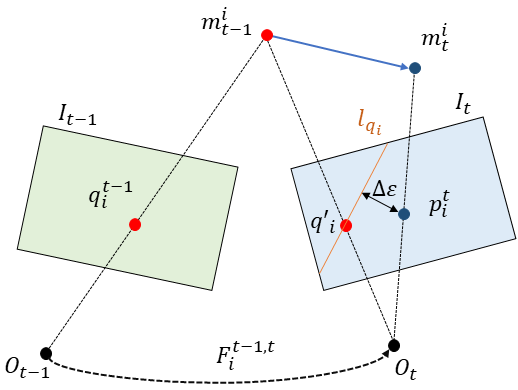}
        \vspace{-0.5em}
        \caption{Illustration of the reprojection error}
    \label{geo_}
\end{figure}

The geometric module, as illustrated in fig \ref{geo_},  uses the epipolar geometry  to link matched keypoints of two distinct images by reprojection constraints.
\vspace{-1.0em}
\vspace{-0.5em}
\newline
The epipolar reprojection constraint is expressed as :
\begin{equation}
l_{q_{i}} = F \cdot (q_{i})^{T} = \left[ \begin{array}{l}
        X_{q_{i}} \\
        Y_{q_{i}}\\
        Z_{q_{i}}
    \end{array}  \right] \label{eq4}
\end{equation}
Where $q_{i}^{t-1}$ is the keypoint of a frame $t-1$ and $p_{i}^{t}$  the associated matched keypoint of the next frame $t$,  $F$ denoting the fundamental matrix describing the camera motion between these two frames and $l_{q_{i}}$ the epipolar line being the reprojection of the keypoint $q_{i}$ in the frame $t$.
The reprojection error $\Delta \varepsilon(p_{i})$ describes the distance between the epipolar line $l_{q_{i}}$ of the keypoint $q_{i}$ and the matched keypoint $p_{i}$ in the current frame. This error is expressed as follows:
\begin{equation}
\Delta \varepsilon(p_{i}) = \frac{| p_{i} F (q_{i})^{T} |}{\sqrt{\Vert X_{q_{i}}\Vert^{2}+\Vert Y_{q_{i}}\Vert^{2}}}    \label{eq5}
\end{equation}
Note that the keypoint $p_{i}$ is considered dynamic if its reprojection error $\Delta \varepsilon(p_{i})$ is greater than an adaptive threshold $\alpha(z_{p_{i}})$ related to the keypoint at the depth $z_{p_{i}}$. 
%This threshold $\alpha(z_{p_{i}}) \in [0.5;0.9]$, inversely proportional to the depth, assumes that the farther a point is, the more its reprojection error represents a large movement due to projective geometry constraints.
Making the threshold $\alpha(z_{p_{i}}) \in [0.5;0.9]$ inversely proportional to the depth is based on the fact that the farther a point is the more its reprojection error corresponds to a large movement due to projective geometry constraints.
% If this error is smaller than the threshold, then it feeds the following normal probability density function to estimate the state probability of the keypoint:
If this error is smaller than the threshold, then it is used into the probability density function to estimate the state probability of the keypoint as follow:
\begin{equation}
P(g_{p_{i}}) = \frac{1}{\sqrt{2\pi}\sigma}\exp\left( -\frac{(\Delta \varepsilon(p_{i}))^{2}}{2\sigma^{2}} \right)   \label{eq6}
\end{equation} 
Where $\sigma$ is set to 1 and represents the standard deviation. 
%This state probability is then scaled to account for the actual dynamic range of the probability.

\subsection{Moving probability update}
The moving probability $P(p_ i)$ represents the state probability either dynamic or static of the keypoint $p_{i}$.
The moving probability $P(p_ i)$ combines the geometric  $P(g_{p_{i}})$ and semantic $P(S_ {p_{i} })$ probability models as follows:
\begin{equation}
P(p_i) = \omega P(g_{p_{i}}) +(1-\omega)P(S_ {p_{i} }) \label{eq7}
\end{equation}
Where $\omega$ is a weight that describes the relevance of the probabilistic model for different keypoint situations.
In uncertain mask areas, $\omega$ is set to 0.5, while in the reliable area, we set it to 0.1.
In the case of a non-matched keypoint included in the mask area, the moving probability $P(p_ i)$ corresponds to the semantic probability $P(S_ {p_{i} })$. Conversely, if it is a matched keypoint out of the mask area, then $P(p_ i)$ takes to the geometric probability $P(g_ {p_{i} })$.
We update the moving probability function using Bayesian filter as expressed: 
\begin{equation}
bel(p_i) = \eta P(\Omega_i | p_i) \int P(p_i | q_i) bel(q_i) dq_i \label{eq8}
\end{equation}
Where $\eta$ is an impact factor to normalize probabilities. 
%Where $\eta = \frac{1}{bel(p_i=d) +bel(p_i=s)}$ as an impact factor to normalize probabilities.
%Initials prior probability $p_0$ and observation likelihood $P(\Omega_i | p_i)$ are set to 0.5. Keypoints with a moving probability greater than 0.5 are considered dynamic points and therefore rejected.

\subsection{Object Interaction Module}
\label{interaction}
\begin{figure}[h]
    \centering
        \includegraphics[width=0.42\textwidth]{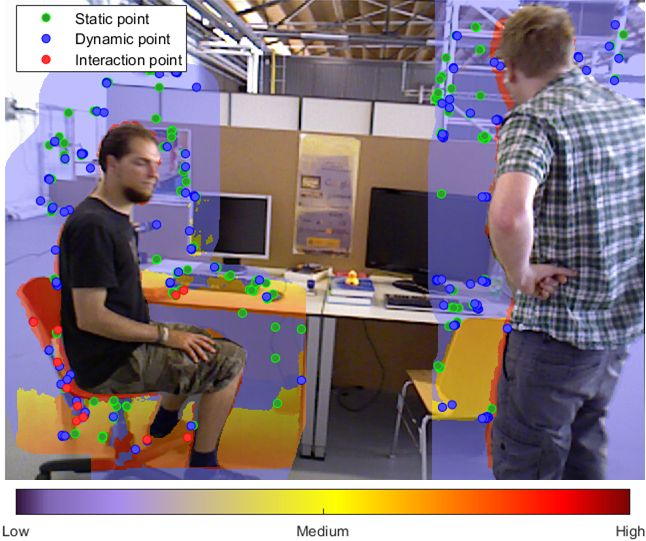}
        \caption{Probability of objects interaction in close areas}
        \small Interaction points are dynamic points obtain through the OIM.
    \label{area}
\end{figure}

The OIM determines possible interactions between humans and inert objects. 
%It implies that the more a point considered static has an image position and a depth close to a human, the more likely it is to interact with. 
This implies that the closer a point, considered as static, is to a human, in terms of 2D position and depth, the more likely it is to interact with it.
Thus, we define an interaction zone related to the human depth (see the colored areas in fig. \ref{area}) in which we estimate the correlation between the point distances in position and depth with the nearest keypoint to humans.   
Consider a static keypoint $p_i$ in this interaction region with a 2D-plus-Depth position smaller than an adaptive threshold.
% Suppose a considered static keypoint $p_i$ in this area has a position and depth distance lesser than an adaptive threshold. 
The OIM examines the dynamics keypoints $p_{id}$ from the geometric module in the neighbourhood of $p_i$ and retains those satisfying the following conditions:
\begin{equation}
 \begin{cases}
    || p_i- p_{id}|| < \delta(z_{p_i}), & \text{$\delta(z_{p_i}) \in [11;48]$}\\
    \Delta(i) = | p_i(z) - p_{id}(z) | < \rho, & \rho = 0.7
  \end{cases} \label{eq9}
\end{equation}
Where $\delta(z_{p_i}) = 48 -4\cdot z_{p_i}$ is an adaptive position distance threshold in pixels, and $\rho$ is a depth distance threshold in meters.
% We calculate the center of gravity $G(p_i)$ of these dynamic points weighting it by its depth distance as follows:
The center of gravity $G(p_i)$ of these dynamic points is estimated using a depth-weighting function given below: 
\begin{equation}
G(p_i) = \frac{\sum_{i=1}^{k} p_{id} \cdot \frac{\rho -\Delta(i)}{\rho}}{\sum_{i=1}^{k} \frac{\rho -\Delta(i)}{\rho}}  \label{eq10}
\end{equation}
The distance $\Delta (G_{pi})$ of this center from the point $p_i$ provides information on the state of $p_i$. A small distance implies that the static point is either very close to these dynamic points or that they encompass it. Thus, the OIM changes the keypoint state to dynamic when $\Delta (G_{pi})$ is smaller than a threshold $\gamma(z_{pi}) \in [10;28]$ inversely proportional to the depth.
\vspace{-1em}

\section{Experiments and results}
\label{Expe}

We have evaluated our method in the fre3 sequence of the public dataset TUM RGB-D dedicated to assessing vSLAM methods in dynamic environments. 
This dedicated sequence provides RGB and depth images, and ground truth trajectories.
It is broken down into two types of sequences, namely sitting (s) and walking (w), which correspond to scenarios with low and high levels of dynamics. 
All experiments were performed on an Intel XENON CPU and Nvidia GPU RTX2080 SUPER. 
%To perform the segmentation, We used the official YOLACT++ model with the backbone Resnet50-FPN pre-trained on the COCO dataset with an image size of 550*550.
Our method has been compared to other state-of-the-art dynamic vSLAM methods in terms of the Absolute Trajectory Error (ATE) and Relative Pose Error (RPE) metrics. 
All results represent the rate of improvement of methods according to the ORB-SLAM3 method performance, where we highlighted in bold the best results of each assessment for better readability.
As shown in Tables \ref{tab_ate}, \ref{tab_rte}, \ref{tab_rte1}, the obtained results clearly show that our method outperform  state-of-the-art in terms of RPE. Whereas, in terms of ATE, we obtain relatively significant improvements for some sequences and slightly below state-of-the-art for others.
Note that the proposed method outperforms the considered methods without requiring extra semantic information and with a lower execution time.
\begin{table}[h]
\caption{Evaluation of the ATE in meters (m).}
\begin{center}
\begin{tabular}{c|cccc}
\hline
\centering{\textbf{Seq.}}&\multicolumn{4}{c}{RMSE}  \\
\cline{2-5}
 \textbf{fr3}& DynaSLAM & DS-SLAM &DP-SLAM& ours \\
\hline
\hline
w/xyz  & \textbf{98.2\%}&97.3\%&97.9\%&97.9\%\\
w/half &\textbf{95.5\%}&95.4\%&94.1\%&94.3\%\\
w/static  &\textbf{98.1\%}&97.8\%&98.0\%&\textbf{98.1\%}\\
w/rpy &\textbf{96.8}\%&60.0\%&94.9\%&96.1\% \\
s/static &-2.0\%&27.7\%&29.8\%&\textbf{33.3\%} \\
\hline
\end{tabular}
\label{tab_ate}
\end{center}
\end{table}
\vspace{-2.5em}
\begin{table}[h]
\caption{Evaluation of the translational RTE in meters (m).}
\begin{center}
\begin{tabular}{c|cccc}
\hline
\centering{\textbf{Seq.}}&\multicolumn{4}{c}{RMSE}  \\
\cline{2-5}
 \textbf{fr3}& DynaSLAM & DS-SLAM &DP-SLAM& ours \\
\hline
\hline
w/xyz  & 94.9\%&92.2\%&95.8\%&\textbf{98.0\%} \\
w/half &91.3\%&90.9\%&61.8\%&\textbf{94.0\%}\\
w/static  &\textbf{98.9\%}&98.7\%&51.8\%&98.1\%\\
w/rpy &89.7\%&65.6\%&54.4\%&\textbf{96.0\%}\\
s/static &-23.5\%&23.5\%&3.8\%&\textbf{40.0\%}\\
\hline
\end{tabular}
\label{tab_rte}
\end{center}
\end{table}
\vspace{-2.5em}
\begin{table}[h]
\caption{Evaluation of the rotational RTE in degree (°).}
\begin{center}
\begin{tabular}{c|cccc}
\hline
\centering{\textbf{Seq.}}&\multicolumn{4}{c}{RMSE}  \\
\cline{2-5}
 \textbf{fr3}& DynaSLAM & DS-SLAM &DP-SLAM& ours \\
\hline
\hline
w/xyz  &92.0\%&89.5\%&40.8\%&\textbf{96.9\%} \\
w/half &89.2\%&88.7\%&55.8\%&\textbf{94.1\%}\\
w/static  &95.7\%&95.5\%&37.1\%&\textbf{96.8\%} \\
w/rpy &88.7\%&65.7\%&34.7\%&\textbf{94.4\%}\\
s/static &-13.6\%&9.0\%&2.4\%&\textbf{27.1}\%\\
\hline
\end{tabular}
\label{tab_rte1}
\end{center}
\end{table}
\vspace{-1.0em}
\newline
We also compared with the DGS-SLAM method. Our method achieves a gain of over 6.6\% ,in terms of ATE, on  fr3 sequence of the RGB-D TUM dataset. From these results we can conclude that the proposed method  is particularly effective for scenarios with high dynamics. 

\section{Conclusion}
\label{conclusion}
Through this study, we have shown the importance of considering the proximity of the keypoints by integrating the depth and distance information concerning the dynamic persons to improve the localization robustness.  
Combining multimodal information allowed us to consider the interaction between humans and inert objects. The results are convincing and show the efficiency of our approach. 
In future work, we will exploit this approach to study other aspects related to process optimization and execution time reduction.
%This allowed us to take into account the interaction that exists between humans and inert objects by combining the information of the geometric module and the proximity information. The results are convincing and show the interest and the relevance of our approach. In this study we have limited ourselves to the analysis of the trajectory estimation. In a future work we will exploit this approach in the study of other aspects related to pose estimation.

%\vfill
%\pagebreak

\bibliographystyle{IEEEbib}
\bibliography{Template}

\end{document}